\documentclass[11pt,a4paper]{article}
\usepackage[hyperref]{acl2020}
\usepackage{times}
\usepackage{latexsym}
\usepackage{graphicx}
\usepackage{booktabs}

\usepackage{microtype}

\aclfinalcopy 
\def\aclpaperid{72} 

\title{Assessing Discourse Relations in Language Generation from GPT-2}

\author{Wei-Jen Ko \\
  Department of Computer Science \\
  The University of Texas at Austin \\
  \texttt{wjko@utexas.edu} \\\And
  Junyi Jessy Li \\
  Department of Linguistics \\
  The University of Texas at Austin \\
  \texttt{jessy@austin.utexas.edu} \\}

\date{}

\begin{document}
\maketitle
\begin{abstract}
Recent advances in NLP have been attributed to the emergence of large-scale pre-trained language models. GPT-2 \cite{gpt2}, in particular, is suited for generation tasks given its left-to-right language modeling objective, yet the linguistic quality of its generated text has largely remain unexplored.
Our work takes a step in understanding GPT-2's outputs in terms of discourse coherence.
We perform a comprehensive study on the validity of explicit discourse relations in GPT-2's outputs under both organic generation and fine-tuned scenarios. Results show GPT-2 does not always generate text containing valid discourse relations; nevertheless, its text is more aligned with human expectation in the fine-tuned scenario. We propose a decoupled strategy to mitigate these problems and highlight the importance of explicitly modeling discourse information.
\end{abstract}

\section{Introduction}

Recent progress in NLP has been marked with the emergence of large-scale pre-trained models, e.g., ELMo~\cite{elmo}, BERT~\cite{bert}, and GPT-2~\cite{gpt2}. Among these, GPT-2 is particularly suitable in natural language generation due to its underlying left-to-right language modeling objective. 
Indeed, GPT-based language models have shown impressive results for open-domain dialogue generation \citep{golovanov2019large,wolf2019transfertransfo,dialogpt}. This has motivated investigations into GPT-2's generated text \cite{see2019massively,wallace-etal-2019-universal}. In particular, using automatic metrics (e.g., cosine similarity, lexical diversity, sentence length), \citet{see2019massively} illustrated that GPT-2 has the ability to generate interesting and coherent text. However, analysis of GPT-2's outputs from deeper linguistic dimensions (e.g., discourse) has largely remained unexplored.

In this paper, we perform the first discourse analysis of GPT-2's outputs, under both organic and fine-tuned scenarios, with the goals of understanding model behavior and pointing towards ways of improvement. We chiefly focus on \textit{discourse relations}, one of the most important linguistic devices for textual coherence. Discourse relations specify the relationships between text spans, for example:
\vspace{-0.3em}
\begin{quote}
    \emph{Jazz is good, \textbf{but} my favorite is country music.}
\end{quote}
\vspace{-0.3em}
The two clauses (also called \emph{arguments}) are connected by a \textsc{contrast} relation, as signaled by the connective \textit{but}. 
Discourse relations are central in establishing textual coherence. For example, they create rhetorical connections between spans in the absence of anaphoric entity mentions \cite{lascarides2008segmented}.
Cognitive experiments have repeatedly shown discourse relations to be highly influential in the mental processing of text \citep{meyer1984effects,horowitz1987rhetorical,millis1993impact,sanders2000role}. 
Spans joined with \textit{incorrect} discourse connectives can seem logically incoherent although they are independently grammatical:
\vspace{-0.3em}
\begin{quote}
    \emph{Jazz is good, \textbf{because} my favorite is country music.}
\end{quote}
\vspace{-0.3em}
The importance of generating good discourse connectives are recognized in prior work in NLG~\cite{biran-mckeown-2015-discourse,callaway-2003-integrating}.

We examine to what extent does GPT-2 generate texts that uphold \textit{plausible} discourse relations, once a discourse connective (usually 1-2 tokens) is generated. 
We present a comprehensive analysis of discourse connectives in both \textit{fine-tuned} generation---specifically, open domain dialogue generation---and \textit{organic} generation directly from GPT-2. We find that GPT-2 generates valid discourse connectives when the relation can be inferred by humans with high agreement, yet struggles to recover less obvious relations.
Our manual analysis reveals the most common connective error is that the relations, signaled by the connectives, \textit{do not hold} between the spans they connect. To this end, we propose a simple remedy: train a connective prediction model and replace incorrect connectives in a post-processing step. This method improves agreement between human and machine-generated connectives in both the fine-tuned and the organic scenarios. Collectively, our results highlight the importance of inferring discourse relations~\cite{xue2015conll}, and explicitly incorporating discourse information in language models~\cite{ji2016latent}, to increase their downstream efficacy.

\section{Experimental Setup}

\paragraph{Fine-tuned.} We choose open-domain dialog generation as our \textit{fine-tuned} scenario. The model conditions on a prompt (dialog turn) and generates a response (next turn). We use the \textsc{PersonaChat} \citep{pc} data for the ConvAI2 challenge. We use 122,499 prompt-response pairs for training and 4,801 pairs for validation.

We fine-tune GPT-2 medium (345M parameters). For compatibility with GPT-2's pre-training, we concatenate the prompt and response (separated by a delimiter) during training. GPT-2 is fine-tuned for 3 epochs using Adam \cite{Kingma2014AdamAM} with a learning rate of 5e-5. The cross-entropy (language modeling) loss is only calculated for the response. At test-time, the model is conditioned on the prompt (and delimiter) and generates the response. Our approach is similar to \citet{dialogpt} and we follow \citet{ko} to encourage generation of informative responses.\footnote{\newcite{ko} used a linguistic metric which performed better than mutual information also used in~\newcite{dialogpt}.}

For decoding, we experimented with both top-$k$ sampling~\cite{fan2018hierarchical} and nucleus sampling~\cite{holtzman2019curious}, and picked the better performing one upon manual inspection of the validation data. We use top-$k$ (k=10) in this scenario.

For quality assurance, we manually evaluate GPT-2's generated responses against SpaceFusion~\cite{spacefusion}, a state-of-the-art RNN-based model, re-trained on \textsc{PersonaChat}. The evaluation is conducted on Amazon Mechanical Turk, where 5 annotators (per HIT) chose between GPT-2 and SpaceFusion responses. GPT-2 (45.5\% chosen) largely outperforms SpaceFusion (16.9\% chosen). For the other 37.7\%, the two are tied.

\paragraph{``Organic'' generation.}
To determine to what extent GPT-2 understands the discourse functions of connectives without the effects of fine-tuning, we engage an \textit{organic} scenario. In this case, we pick out utterances with explicit discourse relations in the dataset, and feed the partial utterance that approximates the first argument of an explicit discourse relation (the part before the discourse connective), along with the connective, into the GPT-2 model; we then let it continue to generate the rest of the utterance. We use \textsc{PersonaChat} to make the results more comparable to the fine-tuned scenario.\footnote{We do not explicitly perform quality assurance for this scenario as we do not fine-tune GPT-2. Details of language modeling performance are discussed in~\newcite{gpt2}.}
We again experimented with both  nucleus sampling and top-$k$, and used nucleus sampling ($p=0.9$) which performed better upon manual inspection.

\section{Assessing explicit discourse relations}
\label{sec:assess}

\begin{table*}
\centering
\small
\begin{tabular}{l|lllllllllll}
\toprule
  &after&and&because&before&but&if&since&so&though&when&while\\
  \midrule
  	\textsc{PersonaChat}&1.4&40.7&4.2&1.1&28.5&4.4&2.8&4.8&1.1&8.8&2.1\\
  	Fine-tuned&0.5&45.7&1.7&0.4&35.9&1.6&2.6&3.7&0.2&5.3&2.4\\
  	Organic&0.5&51.4&4.4&1.0&22.1&5.7&1.5&5.8&0.7&5.1&1.8\\
  \toprule

\end{tabular}
\caption{\% of sentences with a particular discourse connective, of all sentences that contain a connective.}
\label{tab:conndist}
\end{table*}

At a high level, our assessment strategy compares discourse connectives from GPT-2 outputs with human judgment, following existing strategies of discourse relation annotation, which ask annotators to insert connectives between text spans~\cite{pdtb,scholman2017crowdsourcing,yung2019crowdsourcing}. A discourse connective can be considered valid if humans would also insert a connective signaling the same discourse relation when the connective is masked.

\paragraph{Extracting sentences with discourse connectives.} 
We follow prior work~\cite{braud2016learning,ma2019implicit} in the use of heuristics 
to extract sentences with discourse connectives, 
using a list of 11 connectives most frequently observed in \textsc{PersonaChat}: \emph{after, and, because, before, but, if, since, so, though, when, while}.  Specifically, a clause (using verbs as approximations) needs to appear before and after the connective; the connective cannot be immediately followed by a punctuation; and only \emph{and} and \emph{but} can follow a period.  We remove instances of \emph{so} immediately followed by an adjective or adverb. Upon manual inspection of a random sample of 133 extracted sentences, 100\% of them  contain an explicit discourse relation. 

In the \textsc{PersonaChat} training set, $\sim$11\% of the responses contain one of the  connectives. In contrast, the fine-tuned model generates a connective 26\% among all responses, and the organic one  15\%. The increase in percentage is likely because connectives are frequent words in the corpus. Table~\ref{tab:conndist} shows the relative frequencies of these connectives.
Notably, the distribution of connectives is skewed, with \emph{and} and \emph{but} appearing much more often than other connectives, a characteristic similar to other collected examples of discourse relations in the conversation domain~\cite{ma2019implicit}.
\paragraph{Annotating discourse relations.} 
To assess if GPT-2 generate valid discourse connectives, we compare relations signaled by these connectives with relations that humans judge to hold given the rest of the sentence, as in a masked language modeling task. 
Specifically, for each output sentence that contains a discourse connective, we mask the connective\footnote{The workers saw an underlined blank space for the mask. If multiple connectives exist, we only consider the first one in this work.} and show the rest of the sentence to annotators (in the case of dialogue generation, we also show the prompt). They are asked to fill in the blank with a connective that most naturally expresses the relation between the arguments, or \textsc{none} if they think the two segments are not related. This type of insertion is used previously to crowdsource discourse relations~\cite{yung2019crowdsourcing,scholman2017crowdsourcing}. 
To reduce label sparsity, we group the connectives into the four top-level discourse relations in the Penn Discourse Treebank~\cite{pdtb} (\emph{contingency, contrast, expansion, temporal}), and the annotators are asked to choose a group if it contains the connective they think most appropriately fills the blank. To further help annotators, we included unambiguous synonyms of connectives to anchor the relations more. For ambiguous connectives in our list, we put them in all possible relations they signal. The specific groupings are listed below:
\begin{itemize}
\item because, therefore, if, so, since (\textsc{contingency})
\item but, although, though, however, whereas,  while (\textsc{contrast})
\item before, after, when, since, while (\textsc{temporal})
\item and, in addition (\textsc{expansion})
\end{itemize}
We also give the \textsc{none} option if the annotator could not find a suitable connective or that the two text spans are not related.

We use Amazon MechanicalTurk to crowdsource annotations for
1.2K sentences with discourse connectives each for the organic and fine-tuned scenarios. 
Each sentence is annotated by five workers. As quality control, we only allow workers in the US that have completed more than 500 hits with an acceptance rate of $>$98\%. 
\begin{table}[t]
\centering
\small
\begin{tabular}{l|ll}
  \toprule
  &Fine-tuned&Organic\\
  \midrule
  	5&40.9&27.7\\
  	4&27.5&25.0\\
  	3&21.3&30.8\\
   \bottomrule
\end{tabular}
\caption{\% of sentences where the discourse relation is agreed by $n\in \{3,4,5\}$ annotators.}
\label{tab:agree}
\end{table}
\begin{table}[t]
\centering
\small
\begin{tabular}{l|ll}
  \toprule
  &Fine-tuned &Organic\\
  \midrule
  	contingency&6.4&12.5\\
  	temporal&5.1&6.2\\
  	contrast&35.1&27.1\\
  	conjunction&52.5&53.0\\
  	no relation&0.9&1.1\\
   \bottomrule
\end{tabular}
\caption{\% of annotated majority relations.}
\label{tab:reldist}
\end{table}

Table~\ref{tab:agree} shows the percentage of sentences whose discourse relation is agreed upon by 5, 4, and 3 workers; Table~\ref{tab:reldist} shows the frequency distribution of majority relations (one that is agreed by $\ge3$ workers). 
For the fine-tuned case, 89.7\% of the sentences have a majority relation; inter-annotator agreement measured by Krippendorff's alpha is 0.508, indicating moderate agreement~\cite{artstein2008inter}. This shows that in most cases, readers are able to infer a discourse relation between the spans of text given, and they do so consistently.  
Similarly in the organic case,  83.5\% of the sentences have a majority relation. However, relations agreed by $\ge4$ workers are much fewer; Krippendorff's alpha is also at a lower value of 0.382. After adjudicating 70 examples with no majority, 
we find that the cause of lower inter-annotator agreement is likely due to the fact that more than one relation can often hold, and in other cases, the quality of the generated text is low. 

\begin{table}[t]
\centering
\small
\begin{tabular}{l|ll}
  \toprule
  &Fine-tuned &Organic\\
  \midrule
  $\ge$ 3 & 81.5 & 74.9 \\
  \midrule
  	5&94.0&92.6\\
  	4&75.6&79.2\\
  	3&64.3&53.2\\
  \bottomrule
\end{tabular}
\caption{\% of connectives in generated texts that are consistent with human annotation, stratified by the \# of annotators agreeing on the relation.}

\label{tab:genagree}
\end{table}

\begin{figure}[t]
\begin{center}
\includegraphics[width=0.8\columnwidth]{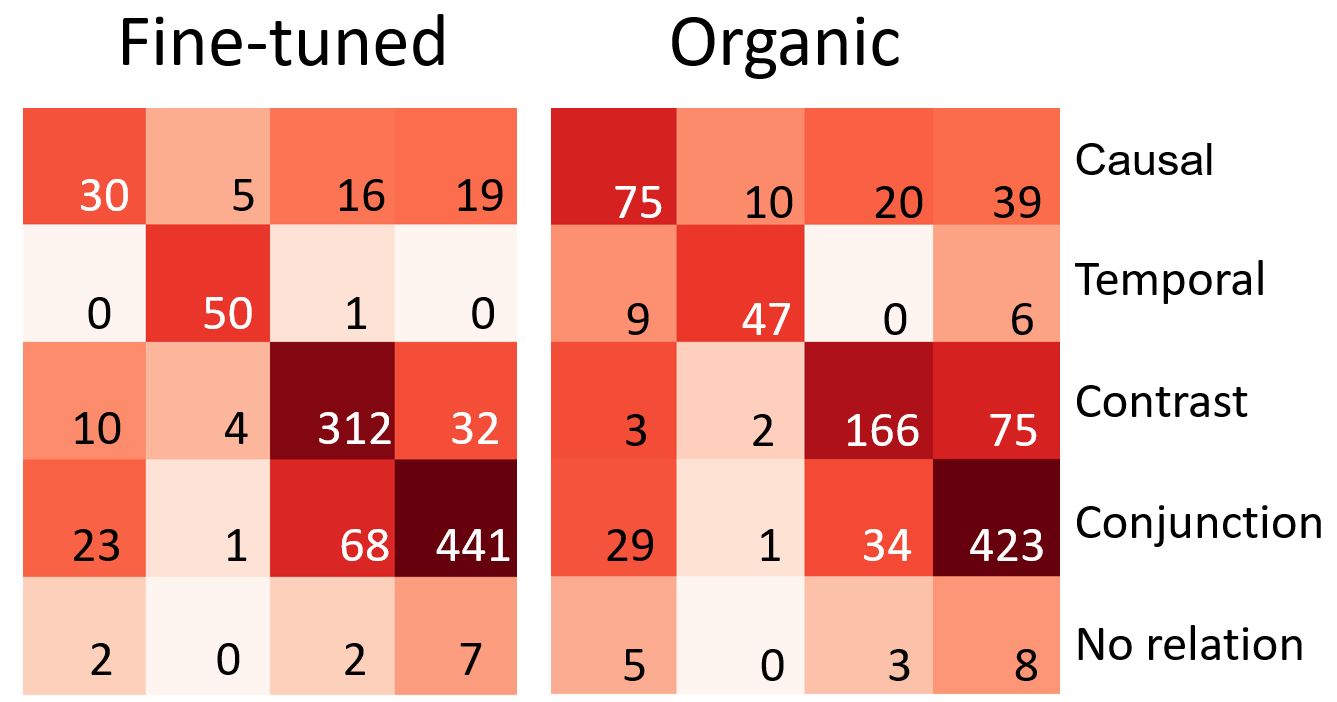}
\end{center}
\caption{Confusion matrix for human labeled relations vs.\ generated connectives (after grouping into relations). Darker color indicates more instances. Vertical axis: human annotated relation. Horizontal axis: GPT-2.}
\label{fig:mat1}
\end{figure}

\paragraph{Assessment results.}
Table~\ref{tab:genagree} shows the percentage of sentences where the connective in the generated text agrees with the majority relation annotated by humans; we also show the results stratified by how many people agree on the relation. For the connectives \emph{since} and \emph{while} which can signal two relations, we count the model as correct if either relation is annotated by humans. The results reveal that a wrong connective could be a prominent source of error in GPT-2 generation, though the fine-tuned model agrees better with humans. Notably, for relations that humans agree more consistently, the models also generate correct relations more often. This hints that GPT-2 captures obvious, unambiguous relations better. Figure~\ref{fig:mat1} shows a confusion matrix comparing human labeled relations (where at least 3 annotators agree) with GPT-2 generated ones.

\section{Fixing discourse connectives}\label{sec:fix}
As a first step to fix erroneous connectives, we propose a  post-processing technique that does not require retraining a model or modifying model structure: replacing generated discourse connectives with ones from a connective prediction model. 
This task is related to discourse relation classification (e.g., \newcite{xue2015conll},
\newcite{nie2019dissent}), yet there is no annotated corpora on the dialog domain. While  \newcite{ma2019implicit} mined discourse relations from conversations, using their data yielded inferior performance in preliminary experiments.

\paragraph{Connective prediction model.}
We train a model to predict the masked discourse connective given the rest of the sentence, or \textsc{none} if no relation. For training, we extract 1 million sentences from Reddit that contain discourse connectives, using the heuristics in Section~\ref{sec:assess}. We restrict the length of sentences to be 7-25 tokens, similar to that in PersonaChat.
The model is fine-tuned on the pre-trained BERT-base-uncased model~\cite{bert}, where the text before the connective is used as sentence A, and text after the connective is used as sentence B. We add an additional classification layer taking the learned \texttt{[CLS]} representation as input. 
To obtain training data for the \textsc{none} class, 
we add 300K synthesized examples with sentence A and sentence B sampled from different posts, approximating the absence of discourse relations. 

The model is fine-tuned for 3 epochs on Reddit using a learning rate of 5e-6. The classification accuracy on the validation set of \textsc{PersonaChat} is 0.743 and macro-F1 is 0.649.
In the organic setting, we directly apply this model to predict the masked connective. In the fine-tuned setting, to obtain a better model in the  domain of \textsc{PersonaChat},
we fine-tune the model for 1 epoch on the training set of \textsc{PersonaChat}. 
The classification accuracy  
improved by 3\% and macro-F1 by 5\%.\footnote{Note that this improvement does not translate to a better model for the organic scenario, since GPT-2's output without fine-tuning does not fall in the \textsc{PersonaChat} domain.}

\begin{table}[t]
\centering
\small
\begin{tabular}{l|ll|ll}
\toprule
  & \multicolumn{2}{c}{Fine-tuned} & \multicolumn{2}{c}{Organic}\\
  \midrule
  & GPT-2 & predicted & GPT-2 & predicted \\
  \midrule
    $\ge$ 4&0.781&0.828* & 0.839&0.883*\\
  	$\ge$ 3&0.760&0.789 & 0.726&0.766*\\
 \bottomrule
\end{tabular}
\caption{Consistency between human annotated and predicted discourse relations, measured in macro-F1 of the four relation types. ($\ge n$): $\ge n$ annotators agree on a relation. (*): $p<0.05$ on a bootstrapping test.}
\label{tab:postpros}
\end{table}
\begin{table}[!h]
\centering
\small
\begin{tabular}{l|ll|ll}
\toprule
  & \multicolumn{2}{c}{Fine-tuned} & \multicolumn{2}{c}{Organic}\\
  \midrule
  & GPT-2 & predicted & GPT-2 & predicted \\
  \midrule
  	$\ge$ 4&86.8&89.2* & 86.6&91.4*\\
  	$\ge$ 3&81.5&82.9 & 74.9&80.4*\\
  	$\ge$ 2&84.1&85.9* & 80.0&84.5*\\
    \bottomrule
\end{tabular}
\caption{Consistency between human annotated and predicted discourse relations, measured in \emph{accuracy}. ($\ge n$): calculated on all sentences that $\ge n$ annotators agree on a relation. (*): $p<0.05$ on a binomial test.}
\label{tab:accuracy}
\end{table}

\paragraph{Post-processing results.}
With this connective prediction model, we replace connectives in generated outputs with the predicted ones. We evaluate whether the predicted connectives align better with human judgments, after collapsing to discourse relation types. We see the \textsc{none} prediction (4.4\% for fine-tuned and 17.5\% for organic) as an indicator that the sentence is not coherent, and resample from the model for a new sentence. These cases are not included in the results.  Appendix~\ref{sec:appendix} shows several examples illustrating connectives in the generated text and those predicted by the classifier. 

Table~\ref{tab:postpros} and Table~\ref{tab:accuracy} show the consistency between a connective in the sentence and its corresponding human labeled discourse relation after post-processing,  measured by macro-F1 and accuracy respectively. We stratify results according to the agreement among human annotators. We also show the accuracy of cases where $\ge2$ annotators agree to account for the possibility of multiple valid relations. For both fine-tuned and organic scenarios, the predicted connective aligns closer to human labels than those generated by GPT-2. 
   
\begin{figure}[t]
\begin{center}
\includegraphics[width=0.8\columnwidth]{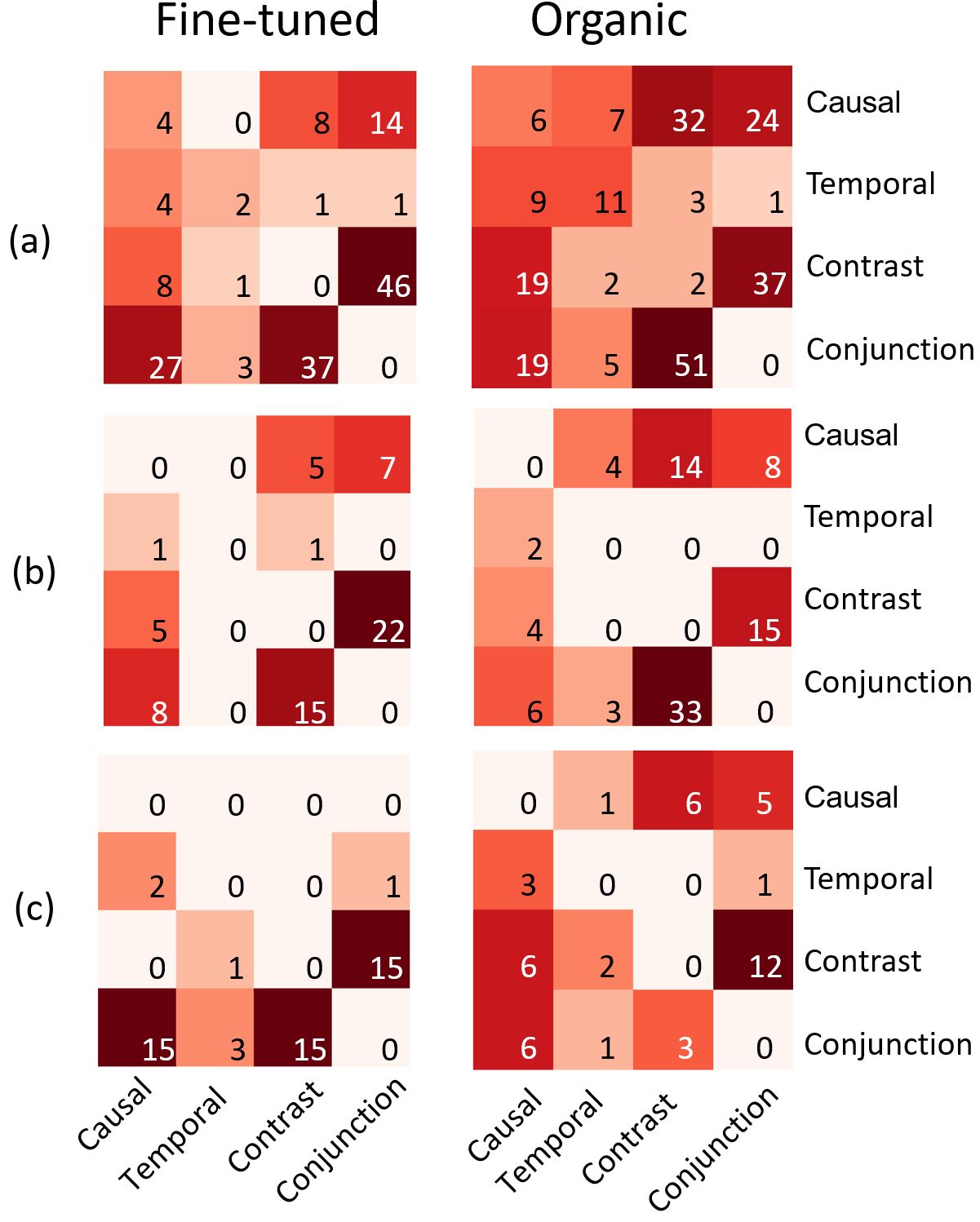}
\end{center}
\caption{Confusion matrix for GPT-2 (vertical axis) vs.\ connective prediction model (horizontal axis). Darker color indicates more instances. (a): all \emph{changed} connectives; (b): sentences that the GPT-2 connectives are inconsistent with human labels, but the connective prediction model gave correct predictions; (c): sentences that the GPT-2 connectives are consistent with human labels, but the connective prediction model gave incorrect predictions. Changed connectives in the same relation class are also included.}
\label{fig:mat2}
\end{figure}

Figure~\ref{fig:mat2} compares the prediction between GPT-2 and the connective predictor for post-processing (Fig.~\ref{fig:mat2}(a)). It illustrates the types of relations that the connective model replaced correctly (Fig.~\ref{fig:mat2}(b)) and incorrectly (Fig.~\ref{fig:mat2}(c)). This shows that the better performance of the model is not due to simply preferring the most frequent class.

The improvement is notably more substantial for the organic case, an indication that fine-tuning GPT-2 nudges the model very close to what the connective prediction model learns. 
The overall improvement is likely due the connective prediction model having access to text before and after the connective, while the initial language generation model does not.  
This finding points to future work on considering stronger discourse-related signals~\cite{ji2016latent} and stronger models for inferring relations.

\section{Conclusion}
This work presents an assessment of discourse relations in organic and fine-tuned language generation from GPT-2. We find that the understanding of discourse connectives are present in these models but are limited, especially when the relation requires more inference. We present a post-processing strategy to replace generated connectives, such that they align better with human expectation.

\section*{Acknowledgements}
This work was partially supported by the NSF Grant IIS-1850153, and an Amazon Alexa Graduate Fellowship. We thank Shrey Desai, Greg Durrett, and the anonymous reviewers for their helpful feedback.

\bibliography{acl2020}
\bibliographystyle{acl_natbib}

\appendix
\clearpage

\section{Example sentences}\label{sec:appendix}
We show several examples below for both fine-tuned and organic scenarios. We list the text that GPT-2 generated (with the connective bolded), and the connective that our classifier predicted (in the subsequent line). 
\subsection{Fine-tuned}
\begin{itemize}

\item \emph{GPT-2}: I do work out at the gym \textbf{but} not as often.\newline 
\emph{Connective classifier}: but\newline
(In this case, GPT-2 produced a plausible connective, and the classifier also predicted the same connective.) 

\item \emph{GPT-2}: My husband is a detective \textbf{so} he loves my family . \newline 
\emph{Connective classifier}: and\newline 
(In this case, GPT-2 did not produce a plausible connective, and the connective classifier was able to correct it.)

\item \emph{GPT-2}: I 'm a housewife , \textbf{but} i also take care of my children\newline 
\emph{Connective classifier}: but\newline 
(In this case, GPT-2 did not produced a plausible connective, neither did the classifier.) 
\end{itemize}

\subsection{Organic}
\begin{itemize}
\item \emph{GPT-2}: It was hard for me to get into college \textbf{and} I 'm still in a wheelchair.\newline 
\emph{Connective classifier}: because\newline 
(In this case, GPT-2 did not produce a plausible connective, and the connective classifier was able to predict a more plausible one.)
\item \emph{GPT-2}: I agree .\ they insist that \textbf{while} they will not pursue civil or criminal action , that they have agreed to withdraw their complaints. \newline 
\emph{Connective classifier}: while\newline 
(In this case, GPT-2 produced a plausible connective, and the classifier also predicted the same connective.) 
\end{itemize}

\end{document}